\documentclass[letterpaper, 10 pt, conference]{ieeeconf}  

\IEEEoverridecommandlockouts                              

\usepackage{graphics} 
\usepackage{epsfig} 
\usepackage{mathptmx} 
\usepackage{times} 
\usepackage{amsmath} 
\usepackage{amssymb}  
\usepackage{mathtools}
\usepackage{svg}
\usepackage{graphicx}      
\usepackage{comment}
\usepackage{xcolor}
\usepackage{subfigure}
\usepackage{cite}
\usepackage{bm}
\usepackage[capitalise, noabbrev]{cleveref}
\usepackage{fixmath}
\usepackage{algorithm}
\usepackage{algpseudocode}
\graphicspath{{pics}}

\newcommand{\Log}{\mathrm{Log}}
\newcommand{\Exp}{\mathrm{Exp}}
\newcommand{\TR}{\textit{TR}}
\newcommand{\PR}{\textit{PR}}

\input{abkuerzung.sty}

\title{\LARGE \bf
Tele-rehabilitation with online skill transfer and adaptation in $\mathbb{R}^3 \times \mathit{S}^3$
}

\author{
Tianle Ni$^{1\dagger}$, Xiao Chen$^{1\dagger}$, Hamid Sadeghian$^{1}$ and Sami Haddadin$^{2}$
\thanks{$^{1}$Authors are with the MIRMI - Munich Institute of Robotics and Machine Intelligence, Technical University of Munich, Germany.
        {\tt\small tianle.ni@tum.de, xiaoyu.chen@tum.de}}%
\thanks{$^{2}$Author is with Mohamed bin Zayed University of Artificial Intelligence
        {\tt\small Sami.Haddadin@mbzuai.ac.ae}}%
\thanks{$^{\dagger}$Equal Contribution.}%
}

\begin{document}

\maketitle
\thispagestyle{empty}
\pagestyle{empty}

\begin{abstract}

This paper proposes a tele-teaching framework for the domain of robot-assisted tele-rehabilitation. The system connects two robotic manipulators on therapist and patient side via bilateral teleoperation, enabling a therapist to remotely demonstrate rehabilitation exercises that are executed by the patient-side robot. A 6-DoF Dynamical Movement Primitives formulation is employed to jointly encode translational and rotational motions in $\mathbb{R}^3 \times \mathit{S}^3$ space, ensuring accurate trajectory reproduction. The framework supports smooth transitions between therapist-led guidance and patient passive training, while allowing adaptive adjustment of motion. Experiments with 7-DoF manipulators demonstrate the feasibility of the approach, highlighting its potential for personalized and remotely supervised rehabilitation. 

\end{abstract}

\section{INTRODUCTION}

The rapid growth of the aging population is accompanied by a sharp rise in the number of stroke survivors and patients with other neurological injuries, such as traumatic brain injury and spinal cord injury. These conditions often leave individuals with long-term motor impairments that limit their ability to perform activities of daily living, highlighting the need for innovative therapeutic interventions to restore lost or impaired motor function. Effective rehabilitation typically requires intensive, repetitive, and prolonged training sessions. However, healthcare systems are increasingly constrained by a shortage of professional therapists, creating a significant imbalance between the rising demand for rehabilitation services and the limited clinical resources available. This mismatch restricts the frequency and intensity of therapy that patients can receive, even though both are critical for functional recovery.

Robot-assisted rehabilitation has emerged as a promising approach to bridge this gap by delivering consistent, repeatable, and quantifiable training while alleviating the burden on therapists \cite{luo2019greedy,Duschau-Wicke2010}. Robots are capable of providing personalized assistance \cite{pezeshki2025personalized} tailored to the patient’s motor intent and performance, thereby playing a crucial role in promoting neuromuscular recovery \cite{Asl2020Field}. Moreover, research shows that rehabilitation outcomes improve when patients are actively engaged in training with minimal external support, as active participation enhances neural plasticity \cite{luo2019greedy}. To encourage this, active rehabilitation strategies have been developed, enabling robotic systems to deliver only the minimum level of support required for successful task execution\cite{luo2019greedy,Hussain2016Gait,Teramae2018EMG,Shi2020Attitude}.

Despite these advances, most robotic rehabilitation platforms\cite{chen2025upper, zhang2025force, huang2024personalized, zhang2020development} remain confined to in-clinic use and lack mechanisms for remote monitoring or supervision. This limitation reduces accessibility, particularly for patients who would benefit from continued therapy at home, underscoring the growing importance of tele-rehabilitation solutions. To address this challenge, a tele-rehabilitation system for the upper extremity using a humanoid robot was proposed in \cite{chen2023tele}. The framework consists of two robotic stations, located at the patient and therapist sides, and employs optimal control and game theory \cite{Pezeshki2023Game}, modeling the patient, therapist, and robot as three agents sharing a common cost function. The system is capable of delivering passive, active, and assist-as-needed rehabilitation through dynamic adaptation of the agents’ roles.

\begin{figure}[!t]
\centering
\includegraphics[width=0.78\textwidth]{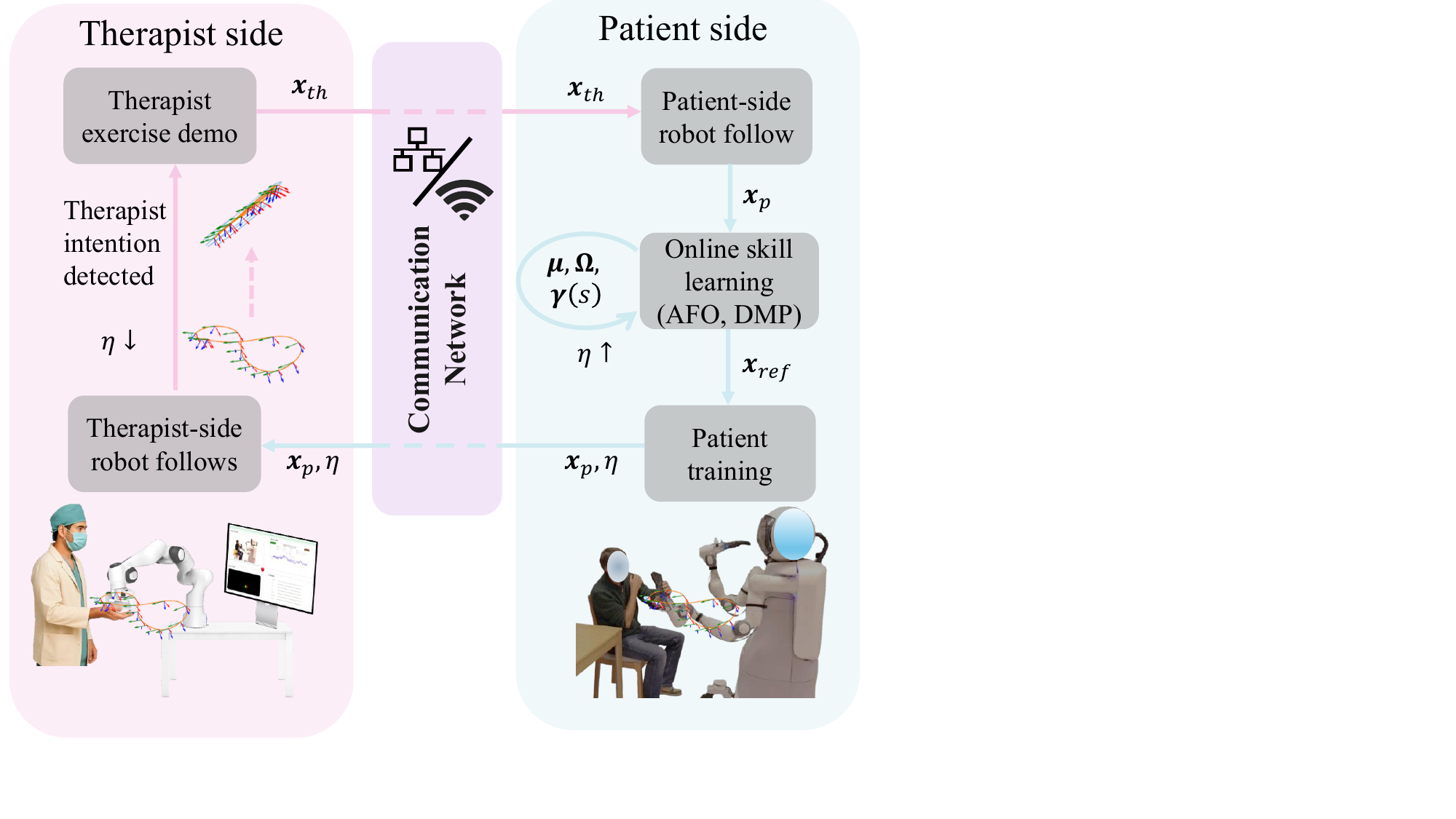}
\vspace{-3em}
\caption{A multi-modal tele-rehabilitation scenario: The patient and robot are in contact through the distal part with an appropriate gamified interface on the screen. On the remote side, the therapist can directly interact with the patient, adjust and observe the parameters and indices, and teach a new exercise through the robot and the provided web-based interface.}
\label{fig:telerehab}
\end{figure}

A therapist-in-the-loop, robot-assisted mirror rehabilitation approach was introduced in \cite{shahbazi2015therapist}, enabling a remote therapist to actively participate in the rehabilitation process. Unlike in clinical settings, where the therapist has direct access to both the patient and the robot, remote supervision imposes significant technical challenges. A transparent robotic interface is required at the therapist’s side to provide high-fidelity interaction with the patient-robot system. Additionally, a robust communication infrastructure is essential to ensure reliable transfer of sensory and audio-visual data. Network issues such as time delays and packet loss can adversely affect both the safety and performance of teleoperation, motivating ongoing research into advanced compensation techniques \cite{sharifi2018patient,chen2024online}.

In this paper, we propose a tele-rehabilitation system that enables remote therapeutic guidance and patient training. The system consists of two synchronized robotic manipulators connected through bilateral teleoperation channels, as illustrated in \cref{fig:telerehab}. To support online and remote rehabilitation, the teleoperation framework is combined with kinesthetic teaching, allowing a therapist to guide exercises in $\mathbb{R}^3 \times \mathit{S}^3$ space in real time. The therapist operates the therapist-side robot—while the patient interacts with it through their own robot—to demonstrate periodic rehabilitation motions. During these demonstrations, the therapist directly guides the patient-side robot through teleoperation, ensuring that the patient follows the prescribed therapeutic motions. The therapist can flexibly adjust motion parameters—including speed, scale, and trajectory shape—based on the patient’s feedback. Once the learning is finished, autonomy is gradually shifted to the patient-side robot so that the impedance controller on this side tracks the reproduced trajectory. The therapist can intervene at any time to introduce new rehabilitation motion profiles in $\mathbb{R}^3 \times \mathit{S}^3$ space, which are encoded and adapted online. This design enables seamless transitions between therapist-led guidance and patient passive training.

The main contributions are as follows:
\begin{itemize}
    \item  A tele-teaching framework that enables remote therapeutic guidance and patient training with periodic rehabilitation motions in $\mathbb{R}^3 \times \mathit{S}^3$ space.
    \item  A dynamic leader-follower bilateral teleoperation control architecture.
    \item  A novel autonomy allocation mechanism based on operator intention and skill learning convergence enables intuitive and seamless transitions between therapist-led guidance and patient passive training.

\end{itemize}

\section{PRELIMINARIES}

\subsection{m-sphere Riemannian manifold}
\label{sec: manifold}
The manifold $\mathit{S}^m=\left\{\boldsymbol{Q} \in \mathbb{R}^{m+1}:\|\boldsymbol{Q}\|=1\right\}$ is a sphere in ${m+1}$ dimensional space. 
 $\mathit{S}^m$ are used to represent directions and orientations in robotic applications. For instance, the robot's End-Effector (EEF) orientation in 3D-space can be described using the space of unit quaternions $\mathit{S}^3$. To encode the robot's rotational motions by quaternion, angular velocity, and angular acceleration, we utilize Dynamical Movement Primitives (DMP)\cite{6797340} in the $\mathit{S}^3$ manifold. We introduce the logarithmic map, exponential map, and trivialization to demonstrate how to operate on the $\mathit{S}^3$ manifold, particularly for applications in DMP.

\subsubsection{Logarithmic map}

Geodesics are the minimum-length curves between two quaternions on the $\mathit{S}^3$ manifold \cite{Calinon25RCFS}.
In \cref{eq:logmap}, the logarithmic map projects the geodesic onto the tangent space at the base quaternion $\boldsymbol{Q}_0$. Intuitively, logarithmic map estimates the difference between two quaternions $\vQ, \vQ_0 \in \mathit{S}^3$ and is defined as, 
\begin{equation}
\delta\boldsymbol{q}
= \Log_{\vQ_0}^{q}\!\left(\vQ\right)
= \frac{\vQ
       - \bigl(\vQ_0^{\top}\vQ\bigr)\,\vQ_0}
       {\bigl\|\vQ
             - \bigl(\vQ_0^{\top}\vQ\bigr)\,
               \vQ_0\bigr\|}
  \, d(\vQ_0,\vQ),
\label{eq:logmap}
\end{equation}
where the geodesic distance is
\[
d(\vQ_0,\vQ)
   \equiv \arccos\!\bigl(\vQ^{\top}\vQ_0\bigr).
\]
The resulted $\delta\boldsymbol{q}$ belongs to $ \mathcal{T}_{\vQ_0} \mathit{S}^3$.

\subsubsection{Exponential map}

The exponential map $\mathcal{T}_{\vQ_0} \mathit{S}^3 \mapsto \mathit{S}^3$ is the inverse of the logarithmic map, which is defined as 
\begin{equation}
\vQ=\operatorname{Exp}_{\vQ_0}^q(\delta\boldsymbol{q})=\vQ_0 \cos (\|\delta\boldsymbol{q}\|)+\frac{\delta\boldsymbol{q}}{\|\delta\boldsymbol{q}\|} \sin (\|\delta\boldsymbol{q}\|).
\label{eq: exp map}
\end{equation}
It projects the point $\delta \boldsymbol{q}$ in the tangent space at the base quaternion $\boldsymbol{Q}_0$ back onto the manifold $\mathit{S}^3$, yielding the quaternion $\boldsymbol{Q}$ so that $\boldsymbol{Q}$ lies on the geodesic starting at $\boldsymbol{Q}_0$ in the direction $\delta \boldsymbol{q}$ \cite{Calinon25RCFS}.

\subsection{Trivialization}
In robotics, the EEF's orientation is represented at the body frame or world frame. What we get from sensors and models are EEF's current orientation expressed as a quaternion, angular velocity $\boldsymbol{\omega}$, and angular acceleration $\dot{\boldsymbol{\omega}}$ with respect to the body frame. In other words, angular velocity is in the tangent space at the identity quaternion $\boldsymbol{\omega} \in \mathcal{T}_{\boldsymbol{1}} \mathit{S}^3$. However, naive time derivative of quaternion $\boldsymbol{\dot{Q}}_t = \Log_{\boldsymbol{Q}_t}^{q}\!\!\Bigl(\boldsymbol{Q}_{t+\delta t}\Bigr)/\delta t$ is in $\mathcal{T}_{\boldsymbol{Q}_{t}} \mathit{S}^3$. To align the sensory information of the angular velocity, we apply trivialization\cite{lee2003smooth} to map data in $\mathcal{T}_{\boldsymbol{Q}_{t}} \mathit{S}^3$ to $\mathcal{T}_{\boldsymbol{1}} \mathit{S}^3$. In this work, we use left trivialization introduced as follows. 

To align with sensory information, we remap $\delta \vq$ from $\mathcal{T}_{\vQ_0} \mathit{S}^3$ to $\mathcal{T}_{\boldsymbol{1}} \mathit{S}^3$ through trivialization\cite{lee2003smooth}. In this work, we use left trivialization introduced as follows. 

Given $\delta \vq \in \mathcal{T}_{\vQ_0} \mathit{S}^3$ from \cref{eq:logmap}, the left translation is defined as $L_{\vQ_0^{-1}} : \delta \vq \mapsto \vQ_0^{-1} \otimes \delta\vq $. The symbol $\otimes$ is the Hamilton product operator. The differential of left translation $d L_{\vQ_0^{-1}}(\delta q)$ maps $\mathcal{T}_{\vQ_0} \mathit{S}^3$ to $\mathcal{T}_{\vQ_0^{-1} \otimes \vQ_0} \mathit{S}^3 \equiv \mathcal{T}_{\boldsymbol{1}} \mathit{S}^3$\cite{lee2003smooth}. The differential of the left translation is identical to “multiply by $\vQ_0^{-1}$” and the group operation is smooth and bilinear in the tangent directions \cite{lee2003smooth}. Therefore, the resulted left trivialization regarding $\vQ_0^{-1}$ is a linear mapping,
\begin{equation}
d L_{\vQ_0^{-1}}(\delta q)=\vQ_0^{-1} \otimes \delta q.
\label{eq:lefttrivialization}
\end{equation}

The logarithmic map projects a quaternion in the $S^3$ sphere onto the tangent space $\mathcal{T}_{\vQ_0}S^{3}$ at the base quaternion $\vQ_0$. The left trivialization with multiplication of $\vQ_0^{-1}$ maps points in $\mathcal{T}_{\vQ_0}S^{3}$ to $\mathcal{T}_{\boldsymbol{1}}S^{3}$. Specifically in robotics, $\mathcal{T}_{\boldsymbol{1}}S^{3}$ corresponds to the body frame. Conversely, the inverse operation of the left trivialization and the exponential map reverses the above process.

The mapping functions between the aforementioned spaces are illustrated below:
\begin{equation}
S^{3} 
\;\xrightleftharpoons[\Exp_{\vQ_0}(\cdot)]{\Log_{\vQ_0}(\cdot)}\;
\mathcal{T}_{\vQ_0}S^{3}
\;\xrightleftharpoons[dL_{\vQ_0}(\cdot)]{dL_{\vQ_0^{-1}}(\cdot)}\;
\mathcal{T}_{\boldsymbol{1}}S^{3}
\end{equation}

\subsection{Operational space description}
\label{sec:oper}
We represent the robot state $\vx$ in $\mathbb{R}^3 \times \mathit{S}^3$ space with the position $\boldsymbol{p}$ and the orientation quaternion $\boldsymbol{Q}$. The difference between the reference robot state $\vx_{ref} = \begin{bmatrix}
    \boldsymbol{p}_{ref}\\
    \boldsymbol{Q}_{ref}
\end{bmatrix}$ and the current robot state is defined using the operator $\ominus$ as $\tilde{\vx} = \vx_{ref} \ominus \vx$, and given by,
\begin{equation}
    \tilde{\vx} = 
    \begin{bmatrix}
        \tilde{\vp} \\
        \tilde{\vq}
    \end{bmatrix}
    =
    \begin{bmatrix}
        \vp_{ref} - \vp \\
        \boldsymbol{e}(\boldsymbol{Q}_{ref},\boldsymbol{Q})
    \end{bmatrix}
    \in \mathbb{R}^6 ,
\end{equation}
where $\boldsymbol{e}(\boldsymbol{Q}_{ref},\boldsymbol{Q})$ represents error between the desired and current quaternion. 

Through logarithmic map in \cref{eq:logmap} and left trivialization in \cref{eq:lefttrivialization}, we can express the error between two quaternions with respect to the body frame at $\mathcal{T}_{\boldsymbol{1}} \mathit{S}^3$.

We define the error as,
\begin{align}
\boldsymbol{e}(\vQ_{ref}, \vQ)&= \operatorname{vec}\!\bigl(
        \vQ_{ref}^{-1} \otimes \Log_{\vQ_{ref}}^{q}\!\left(\vQ\right)\!\bigr) \\
   &= \operatorname{vec}\!\bigl(
        \vQ_{ref}^{-1} \otimes \delta\boldsymbol{q}
     \bigr),
\end{align}

\section{Methodology}

The architecture of the proposed approach is based on leader–follower bilateral teleoperation control, as illustrated in \cref{fig:telerehab}. The therapist-side robot (\textit{TR}) leads the patient-side robot (\textit{PR}) during the therapist’s demonstration phase. The system progressively learns the motion, and when it converges, the \textit{PR} becomes fully autonomous and leads the \textit{TR} to provide feedback to the therapist.
Rehabilitation exercises incorporate both translational and rotational movements, with translational components modeled in Euclidean space $\mathbb{R}^3$ and rotational components on the unit sphere $\mathit{S}^3$. To encode these motions, the periodic DMP framework is employed, iteratively updating its weights until the error between the demonstration and the reproduced trajectory converges. As the \textit{PR} incrementally learns the periodic motions, its autonomy increases, enabling it to reproduce the combined translational and rotational trajectories. This framework allows the therapist to interactively demonstrate and adapt rehabilitation skills online through the robotic interface.

 In the following subsections, three main elements of the proposed approach are introduced. 

\subsection{Teleoperation Control Architecture}

The gravity-compensated dynamics of the therapist robot in Cartesian space ${\vx}_{th}$  is given as,
\begin{equation}
    \vM_{c,th}(\vq_{th})\Ddot{\vx}_{th} + \vC_{c,{th}}(\vq_{th}, \Dot{\vq}_{th}) \Dot{\vx}_{th} = \eta \vu_{th} + \vf_{h,{th}}, 
\end{equation}
where $\vM_{c,th} \in \mathbb{R}^{6 \times 6}$ represents the inertia matrix of the robot as a function of joint space $\vq_{th}$, and $\vC_{c,th} \in \mathbb{R}^{6 \times 6}$ denotes the Coriolis/centrifugal matrix. 
The vector $\boldsymbol{u}_{th} \in \mathbb{R}^6$ is the control command of the \TR{},
and $\boldsymbol{f}_{h,th} \in \mathbb{R}^6$ is the force that the therapist applies to the \textit{TR} by hand. The scalar $\eta \in [0, 1]$ is the level of autonomy for low-level controller as defined in Section \ref{sec:allocation}.

The control command for \TR{} is a PD controller that tracks the robot state from \PR,
\begin{equation}
    \vu_{th} = \vK_{th} \Tilde{\vx}_{th} + \vD_{th} \Dot{\Tilde{\vx}}_{th},
\end{equation}
where $\vK_{th}, \vD_{th} \in \mathbb{R}^{6 \times 6}$ are the positive definite stiffness and damping matrix, respectively, and
$ \Tilde{\vx}_{th} = \vx_{p} \ominus \vx_{th}$ with $\vx_{p}$ as the Cartesian pose of \PR{} transferred to therapist side over communication channel.
The operator $\ominus$ is defined in Section \ref{sec:oper} to calculate the difference between two robot states.
The \TR{} is compliant during the demonstration ($\eta = 0$) but follows the motion of the \PR{} from high-level motion generator introduced in \cref{sec:motion} during the rehabilitation training ($\eta = 1$).

The \PR{} dynamics in Cartesian space is given as,
\begin{equation}
    \vM_{c,p}(\vq_p)\Ddot{\vx}_p + \vC_{c,p}(\vq_p, \Dot{\vq}_p)\dot{\vx}_p = \vu_p + \vf_{h,p},
\label{eq:SR_dy}
\end{equation}
with similar notations as for \TR,  substituting the subscript $-_{th}$ to $-_p$ to represent the \PR.

The control command for \PR{} is a linear blending of the local impedance controller and the remote motion tracking controller, and the transition between the two control commands are realized through the autonomy parameter $\eta$ as, 
\begin{equation}
    \vu_p = \eta \vu_{imp,p} + (1-\eta) \vu_{th,p}.
\end{equation}
The impedance controller for tracking local reference motion is given by,
\begin{equation}
    \vu_{imp,p} = \vM_{c,p} \Ddot{\vx}_{ref,p} + \vC_{c,p} \Dot{\vx}_{ref,p} + \vK_p \Tilde{\vx}_p + \vD_p \Dot{\Tilde{\vx}}_p,
\end{equation}
where $ \Tilde{\vx}_p = \vx_{ref,p} \ominus \vx_p$ with the reference trajectory $\vx_{ref,p}$  generated as described in Section \ref{sec:motion}.

To follow the \TR{} motion, a PD controller is utilized as,
\begin{equation}
    \vu_{th,p} = \vK_{th,p} \Tilde{\vx}_{th,p} + \vD_{th,p} \Dot{\Tilde{\vx}}_{th,p},
\end{equation}
with $ \Tilde{\vx}_{th,p} = \vx_{th} \ominus \vx_p$, to follow the Cartesian pose of the therapist $\vx_{th}$, which is transferred to the patient side over the communication channel.
The \PR{} tracks the \TR{} motion during the demonstration, and gradually switches to track DMP's output trajectories while learning process converges. 

The above architecture demands a reliable communication channel to transfer the pose of both robots bilaterally. Thus, a position-position (PP) structure of teleoperation is implemented here to ensure the synchronization of the \TR{} and \PR.

\subsection{Motion generation}
\label{sec:motion}
The reference rehabilitation skill consists of both translational and rotational motion profiles, where translation is modeled in Euclidean space $\mathbb{R}^3$ and rotation is represented on the manifold $\mathit{S}^3$ using quaternions, owing to their continuous and singularity-free representation. While Euclidean space and $\mathit{S}^3$ share a similar conceptual role, they differ in their algebra: subtraction and addition in $\mathbb{R}^3$ are linear, whereas on $\mathit{S}^3$ they are defined through logarithmic and exponential maps introduced in \cref{sec: manifold}. 

To encode and learn from demonstrations of both translational and rotational motion profiles, periodic DMP expressed in a general six-dimensional representation with Recursive Least Squares (RLS) is employed.  We utilize periodic DMP \cite{6797340} in Euclidean space to encode translational motions and Riemannian periodic DMP \cite{abu2024learning} to encode rotational motions. As periodic DMP and the Riemannian periodic DMP share the same structure, we combine them together in one formulation by introducing a specific operational space.

The rehabilitation skill is encoded in the general periodic DMP in Cartesian space and thus generate the reference trajectory as,
\begin{align}
    \ddot{\boldsymbol{x}}_{ref} &=
\Omega^2\Big(\alpha_z\big(\beta_z\,(\vg_z \ominus \vx_{ref}))-\frac{\dot{\boldsymbol{x}}_{ref}}{\Omega}\big)
+ \boldsymbol{\gamma}(s)\Big), \\
\dot{s} &= \Omega
\label{eq：dmp}
\end{align}
where the change rate of the canonical phase $s$ is the learned frequency $\Omega$ through Adaptive Frequency Oscillator (AFO)\cite{1041514,4762850,Dimeas2020}. The robot state $\boldsymbol{x}_{ref}$ is introduced in \cref{sec:oper}. Here $\boldsymbol{g}_z=(\boldsymbol{g},\boldsymbol{Q}_g)$ is the translational and rotational components of the goal point.

The difference between the goal quaternion $\boldsymbol{Q}_g$ and reference quaternion $\boldsymbol{Q}_{ref}$ in the body frame is

\begin{equation}
    \boldsymbol{e}(\boldsymbol{Q}_g, \boldsymbol{Q}_{ref})= \operatorname{vec}\!\bigl(
        \boldsymbol{Q}_g^{-1} \otimes \Log_{\boldsymbol{Q}_g}^{q}\!\left(\boldsymbol{Q}_{ref}\right)\!\bigr) 
\end{equation}

Similarly, the time derivative of the quaternion in $\mathcal{T}_{\boldsymbol{1}} \mathit{S}^3$ is defined as 
\begin{equation}
    \dot{\boldsymbol{Q}}= \frac{{Q}_t^{-1} \otimes \Log_{\boldsymbol{Q}_t}^{q}\!\!\Bigl(\boldsymbol{Q}_{t+\delta t}\Bigr)} {\delta t} \in \mathcal{T}_{\boldsymbol{1}} \mathit{S}^3, 
\label{eq: Qdot}
\end{equation}
where the corresponding angular velocity is the vector part of $\dot{\boldsymbol{Q}}_t$: $\operatorname{vec}\left(\dot{\boldsymbol{Q}}_t\right) = \tfrac{1}{2}\boldsymbol{\omega}$, and the scalar part of $\dot{\boldsymbol{Q}}_t$ is zero. Thus the time derivative of robot state is,
\begin{equation}
    \dot{\boldsymbol{x}}=\begin{bmatrix}
\dot{\boldsymbol{p}}\\
vec(\dot{\boldsymbol{Q}})
\end{bmatrix}=\begin{bmatrix}
\boldsymbol{v}\\
\tfrac{1}{2}\boldsymbol{\omega}
\end{bmatrix} \in \mathbb{R}^{6}.
\end{equation}
Here, we align the time derivative of the quaternion with the body frame's angular velocity.

Conversely, we apply the exponential map and left trivialization to obtain the next quaternion from the current quaternion and angular velocity generated by DMP,
\begin{equation}
\boldsymbol{Q}_{t+\delta t}
    = \Exp_{\vQ_t}^{q}\!\Bigl(\delta \vq \Bigr)
    = \boldsymbol{Q}_{t}
      \otimes
      \Exp_{\boldsymbol{1}}^{q}\!\Bigl(
        \tfrac{1}{2}\,[0,\boldsymbol{\omega}_{t}]\,\delta t
      \Bigr) ,
\label{eq:quat_update}
\end{equation}

In \cref{eq：dmp}, the forcing term, which is modeled as a normalized weighted sum of basis functions. 
\begin{align}
    \bm{\gamma}(s) &= r \frac{\sum_{i=1}^{N} \bm{\mathrm{w}}_{i} \psi_{i}(s)}{\sum_{i=1}^{N}\psi_{i}(s)} \, , \\
    \psi_{i}(s) &= \exp\left( h(\cos(s -c_{i})-1) \right), \, i = 1, 2, \cdots, N,
\label{eq: forcing}
\end{align}
decides the shape of DMP. $c_{i}$ represents the center of radial basis function (RBF), $h$ denotes the kernel width, and $r$ is the amplitude modulation parameter which is set to $1$. $N$ is the number of RBF. Each vector $\vw_{i} \in \mathbb{R}^6$ of the weights matrix $\vw \in \mathbb{R}^{6 \times N}$ is updated with the Recursive Least Square (RLS) algorithm \cite{Gams2009}. In practice, different number of RBF functions and width for translational and rotational components can be chosen. The forgetting factor $\lambda_{fg}$ in RLS decides how much recent data is used for updating the weights\cite{vahidi2005recursive}. The larger the forgetting factor, the more recent the data is used, but the slower the convergence.

The target forcing term is acquired from therapist demonstrations in $\mathbb{R}^3 \times \mathit{S}^3$ space: $\vx_{th,p}, \dot{\vx}_{th,p}, \ddot{\vx}_{th,p}$ and is assembled by the DMP dynamics introduced in \cref{eq：dmp}

\begin{equation} 
\boldsymbol{\gamma}_{d} = \frac{\Ddot{\vx}_{th,p}}{\Omega^2} - \alpha_z(\beta_z(\vg_z \ominus \vx_{th,p}) - \frac{\dot{\vx}_{th,p}}{\Omega}). 
\label{eq: forcing target} 
\end{equation}

The goal of RLS is to minimize the error $\boldsymbol{err}_{i}(s)$ between the target forcing term from demonstrations and the weights \cite{6797340}, defined as,
\begin{equation}
\boldsymbol{err}_{i}(s) = (1-\mu)\bigl(\boldsymbol{\gamma}_{d}(s)-\boldsymbol{w}_{i}\bigr).
\label{eq:dmp_error}
\end{equation}
The factor $(1-\mu)$ controls the RLS learning rate and gradually decreases to zero once the learning error falls below a specified lower bound. We require that $(1-\mu) = 0$ when the \textit{PR} executes autonomously, thereby preventing further updates to the weight matrix. The design of $\mu$ to meet this requirement is presented in \cref{sec:allocation}.

\subsection{Autonomy Allocation}
\label{sec:allocation}

Autonomy allocation manages the transition between therapist intervention and robot autonomy based on two factors:
(i) whether the skill has been successfully learned and
(ii) whether the therapist intends to end the current demonstration or teach a new skill.
The first factor is assessed primarily from the DMP learning error, and the second is inferred from the measured external forces and torques (wrench) applied to the robot \cite{chen2024online, dalle2024passivity}.
When the system determines that the skill has been learned and no further demonstrations are intended, it transits to autonomous execution; otherwise, it remains in human demonstration mode.
The resulting two-factor variable is defined as the autonomy-level parameter $\eta \in [0,1]$, where $\eta = 0$ corresponds to human demonstration mode and $\eta = 1$ to autonomous mode.

The basic design goal is that, as the autonomy level $\eta$ increases, the learning rates in the DMP and AFO decrease to zero. The following happen simultaneously: 1) switch off optimization and indicating convergence to the current optimum gradually. 2) the controller transitions from a compliant mode to autonomous control by increasing stiffness and impedance \cite{Dimeas2020,9981728}, \cite{chen2024online}.
Thus, the autonomy level simultaneously schedules the learning rate and modulates the low-level controller.
However, in practice, increasing stiffness while decreasing the learning rate can distort the ongoing demonstration, because the robot becomes progressively less compliant, making it difficult for the operator to determine when to conclude a demonstration.

\begin{algorithm}[t]

\caption{Online skill learning and adaptation}\label{alg:cap}
\begin{algorithmic}
\State Input: $\vx_{th}, \dot{\vx}_{th}, \ddot{\vx}_{th}$
\State Output: $\ddot{\vx}_{ref}, \dot{\vx}_{ref}, \vx_{ref}$
\While{running}
  \While{$\mu < 1$}
    \State $\eta \gets 0$
    \State $\boldsymbol{\Omega} \gets AdaptiveFrequencyOscillators(\vx_{p}, \mu)$
    \State $\boldsymbol{\gamma} \gets SkillLearning(\vx_p, \dot{\vx}_p, \ddot{\vx}_p, \boldsymbol{\Omega}, \mu)$
    \State $\ddot{\vx}_{ref}, \dot{\vx}_{ref}, \vx_{ref} \gets MotionGeneration(\boldsymbol{\gamma})$
    \State $\mu \gets LearningAllocation(\tilde{\vx}_p)$
  \EndWhile
  \State $\eta \gets AutonomyAllocation(\vf_{h,th})$
  \State $\mu \gets StatusObservation(\tilde{\vx}_p)$
  \State $\vK \gets \eta \, \vK_{0}$,\quad $\vD \gets \eta \, \vD_{0}$  \Comment{variable stiffness}
  \State $\vu_{imp,p} \gets \vK(\vx_{ref}-\vx_p)+\vD(\dot{\vx}_{ref}-\dot{\vx}_p)$  \Comment{PR tracks $\vx_{ref}$}
\EndWhile
\end{algorithmic}
\end{algorithm}

To address this issue, we decouple the learning rate from the autonomy level while maintaining a similar overall design.
A learning level $\mu \in [0,1]$ to control the skill learning rate, and a low-level autonomy parameter $\eta \in [0,1]$ to modulate the low-level controller are introduced.
Only the parameter $\eta$ affects the teleoperation with variable stiffness and impedance.
Furthermore, $\eta$ is allowed to increase only after the learning level has converged to one ($\mu = 1$), thereby preventing premature stiffness increases from corrupting demonstrations while learning is still in progress.

We define $\mu$ as learning level and $\eta$ as low-level control autonomy level. The rate of change of the learning level is obtained by:
\begin{equation}
    \Dot{\mu} = \begin{cases}
        \max \{\mu_r, 0\}, & \mu = 0\\
        \mu_r, & 0 < \mu < 1\\
        \min \{\mu_r, 0\}, & \mu = 1,
    \end{cases}
\label{eta1}
\end{equation}
with learning autonomy dynamics as, 
\begin{equation}
    \mu_{r} = \left(\frac{\mu}{\rho} + \epsilon \right) \left( 1 - I_s\right),  
\label{eta2 learning}
\end{equation}
To prevent getting stuck at zero, a small parameter $\epsilon$ is introduced while  $\rho$  acts as a scaling parameter.
The component
\begin{equation}
    I_s = \left( \frac{\parallel \vx_{ref} \ominus \vx_{th,p} \parallel}{\lambda_{err}} \right)^4,
\end{equation}
where $\lambda_{err}$ is the hyperparameter serving as tolerance of learning error. The parameter $I_s$ is a value indicates the confidence of the learned skill. When the error decreases below $\lambda_{err}$, the term $(1 - I_s)$ becomes a positive number, which will gradually drive $\mu$ to $1$ and lower the learning rate in \cref{eq:dmp_error} until zero. When the error is larger than $\lambda_{err}$, the $(1 - I_s)$ becomes negative and force $\mu$ to zero, which increases the learning rate to one. 

With the requirement that the low-level control autonomy level $\eta$ can rise only when the learning autonomy level satisfies $\mu = 1$, indicating that the learning process has converged. The rate of change of the low-level control autonomy level is defined as

\begin{equation}
\dot{\eta} =
\begin{cases}
\max\!\bigl\{ \eta_{r}\,\boldsymbol{1}_{\{\mu=1\}},\,0 \bigr\}, & \eta = 0,\\[6pt]
\eta_r^{-} + \eta_r^{+}\,\boldsymbol{1}_{\{\mu=1\}}, & 0 < \eta < 1,\\[6pt]
\min\!\bigl\{ \eta_{r},0 \bigr\}, & \eta = 1,
\end{cases}
\label{eq:eta_gate}
\end{equation}
\[
\eta_r^{+} := \max\{\eta_r,0\}, \qquad
\eta_r^{-} := \min\{\eta_r,0\}, \qquad
\]
where $\boldsymbol{1}_{\{\mu=1\}}$ is one when $\mu=1$, otherwise is zero. This way, we allow rise of low-level autonomy level only when $\mu=1$. The parameter $\eta_{r}$ is given by,

\begin{equation}
    \eta_{r} = \left(\frac{\eta}{\rho} + \epsilon \right) \left( 1 - I_h\right).  
\label{eta2}
\end{equation}
The parameter $I_h$ increases during human intervention for learning a new skill and given by,
\begin{equation}
    I_h = \left( \frac{ \parallel \vf_{h,th}^f \parallel }{\lambda_{\textit{f}}} \right)^4 + \left( \frac{ \parallel \boldsymbol{f}_{h,th}^m \parallel }{\lambda_{\textit{m}}} \right)^4 ,
\end{equation}
considering the interaction force consists of force and moment part, $\boldsymbol{f}_{h,th} = \begin{bmatrix}
    \boldsymbol{f}_{h,th}^f  \\ \boldsymbol{f}_{h,th    }^m
\end{bmatrix} \in \mathbb{R}^6$. The constant parameters $\lambda_{f}, \lambda_{m}$ are chosen for proper interaction of force and moment threshold. Thus, by applying external force from the human demonstrator, the level of autonomy decreases to comply with the human. 

The skill learning and adaptation procedure is summarized in Algorithm \ref{alg:cap}.
\section{Experimental Evaluations}

\begin{figure}[!t]
    \centering
\includegraphics[width=0.4\textwidth]{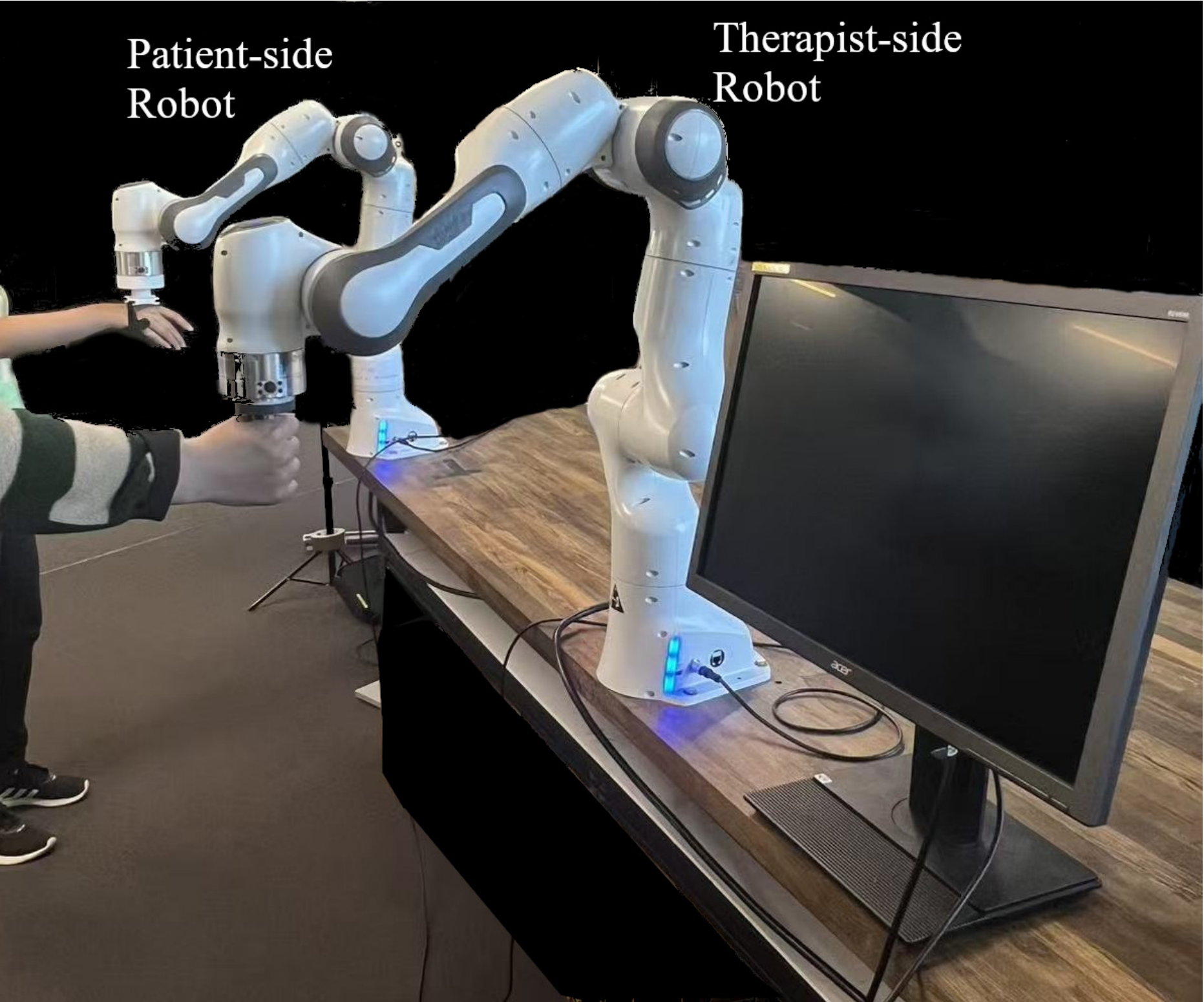}
    \caption{The experiment setup. The user (Therapist) demonstrate the rehabilitation skill on the \TR{}, and the \PR{} holds the patient's hand.}
\label{fig:exp_setup}
\end{figure}

\subsection{Robot Experiment}
The experimental setup consists of two 7 Degrees of Freedom (DoF) Franka Panda robots, as depicted in \cref{fig:exp_setup}. The therapist operates on TR to demonstrate rehabilitation skills, where the patient binds the hand on PR end-effector to receive  the rehabilitation exercise. We assume that the patient remains passive during the exercise. The data between \textit{TR} and \textit{PR} is transmitted through User Datagram Protocol (UDP) with negligible communication delay.

The rehabilitation skills are composed of both translational and rotational motions. A series of different rehabilitation skills in $\mathbb{R}^3 \times \mathit{S}^3$ space are performed continuously to illustrate the learning and adaptation capability across all the dimensions of our tele-rehabilitation framework.

The experiment encompasses the demonstration of five periodic rehabilitation skills denoted as $R_1, \cdots, R_5$ with translational $XYZ$ in Euclidean space and orientational motion expressed by a quaternion. Specifically, the skills contain,
\begin{itemize}
    \item $R_1$: push-and-pull movements along the $X$–axis with rotation about the $Y$ axis,
    \item $R_2$: wave movements along the $Y$ axis with rotation about the $X$ axis,
    \item $R_3$: up-and-down movements along the $Z$ axis with rotation about the $Y$ axis,
    \item $R_4$: an $\infty$ shape along the $Y$ axis with rotation about all axes,
    \item $R_5$: an $\infty$ shape along the $X$ axis with rotation about all axes.
\end{itemize} 

\begin{figure}[!t]
    \centering
\includegraphics[width=0.5\textwidth]{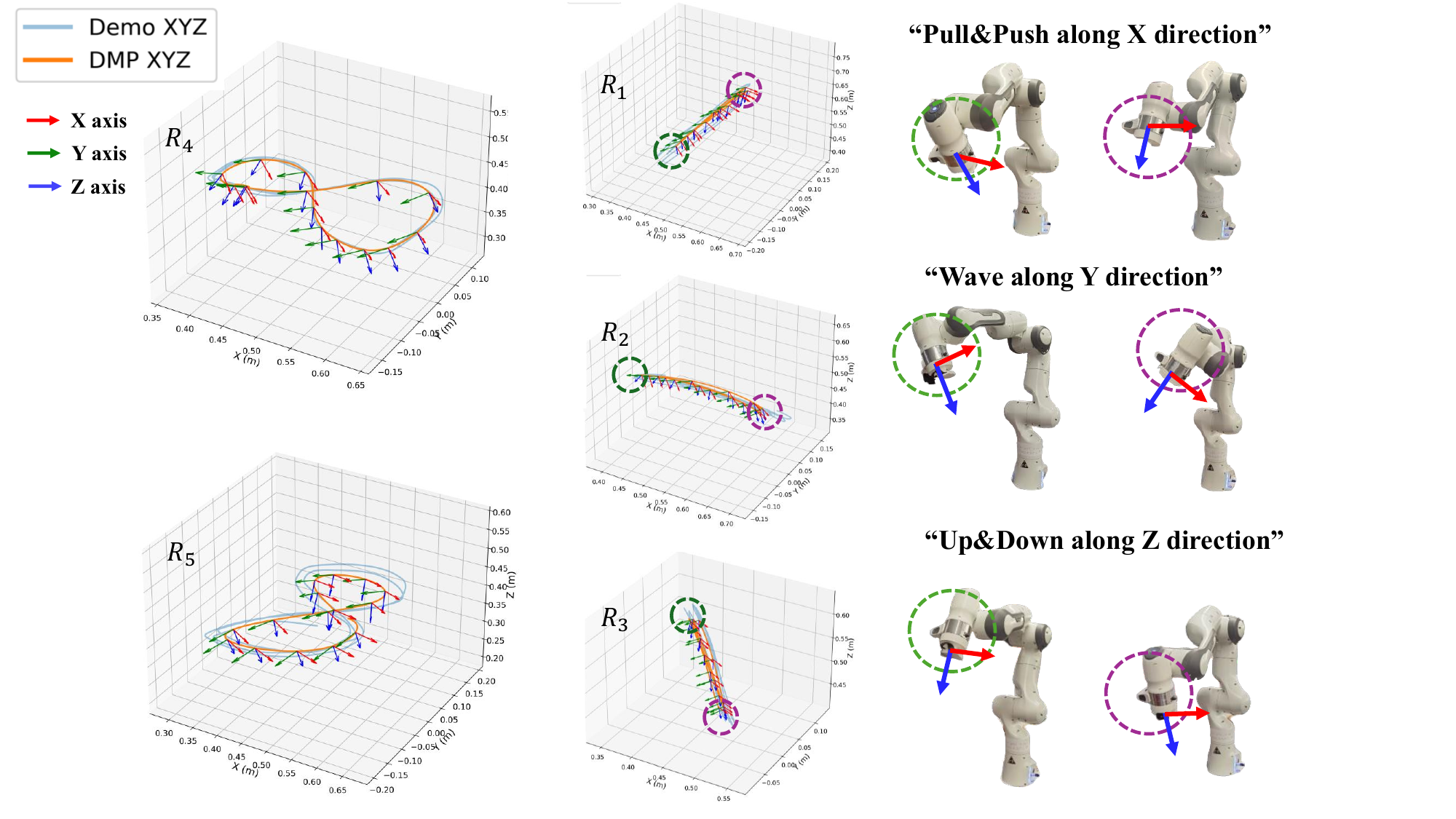}
    \vspace{-2em}
    \caption{Five sample motions with both translational and rotational parts. The blue trajectories are the therapist's demonstrations, and the orange trajectories with the orientation axis attached are the DMP trajectories learned online from demonstrations. For the "line" motions, the corresponding robot poses at the farthest points are marked.}
    \label{fig:rehab_3d_motions}
\end{figure}

To further show the online learning and adaptation capability along all the translational and rotational axes, we conduct an experimental scenario to show the loop of learning-execution-adaption as follows. The therapist demonstrates a skill, the patient follows, and DMP updates its weight to minimize the error between demonstrations and reproduction. Note that the update frequency is 500Hz, while the robot's control loop is 1000Hz. After a few iterations, the current skill is being learned, and the patient trains on the skill passively. After a few rounds, the therapist switches to a new skill demonstration, and DMP adapts to it online, and repeats the previous procedures. The experiment includes the demonstration of three sample periodic  skills $R_5, R_1, R_2$. 

\begin{figure}[!tbh]
    \centering
    \includegraphics[width=0.47\textwidth]{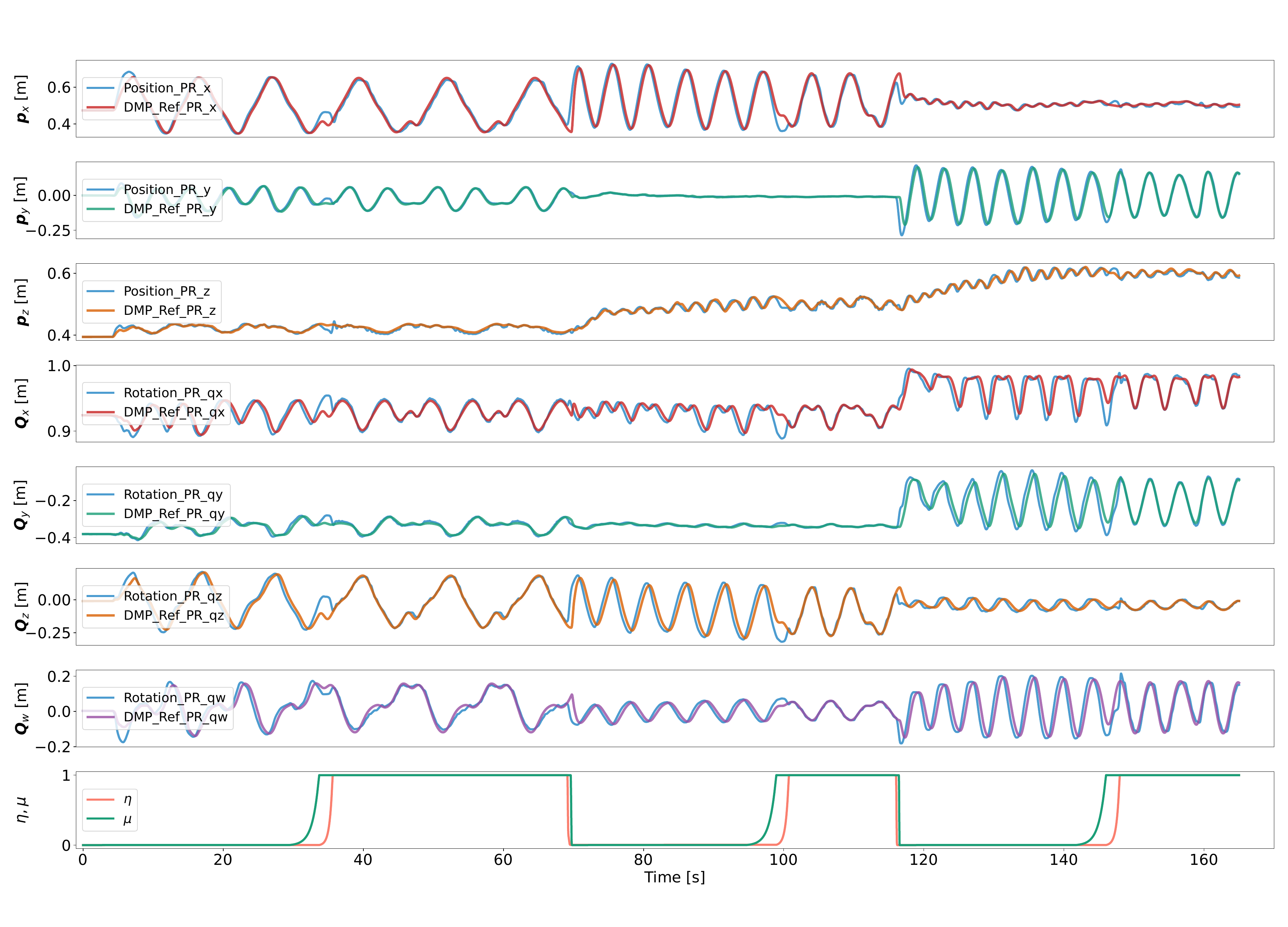}
    \vspace{-1em}
    \caption{Experimental results of the proposed tele-rehabilitation approach. A therapist demonstrates and modifies different periodic rehabilitation skills to the \PR{} through \TR. a), b) and c) include \textit{PR}'s end-effector position $\vp$ and DMP reference trajectory $\vp^{des}$, d), e), f) and g)  \textit{PR}'s end-effector position $\vQ$ and DMP reference trajectory $\vQ_{des}$, h) Autonomy level $\eta$ and learning level $\mu$. Convergence of $\mu$ to one indicates learning is finished. Once learning is finished with no further demonstration, $\eta$ rises to one gradually to switch to the autonomous mode. }
    \label{fig:rehab_online}
\end{figure}

The outcomes of the demonstrated tasks are depicted in \cref{fig:rehab_online}.  
The first demonstration for $R_5$ is started at around $t=5s$. After three iterations (around $t=30 s$), the reference skill converges to the therapist's demonstration in $\mathbb{R}^3 \times \mathit{S}^3$ space, which suggests high confidence in learning. The convergence of the signal shape and frequency increases the learning level $\mu$. Convergence of $\mu$ and low external wrench applied by the therapist increases the low-level autonomy level $\eta$.  The therapist can feel that the robot becomes less compliant, indicating the acquisition of the new skill. After two more iterations at $t=36s$, \textit{PR} gains full autonomy and continuously reproduce the learned rehabilitation skill. At around $t=70s$, the therapist initiates modifications to the previous skill, increasing the external force $\vf_{h,th}$ on \textit{TR}, and the tracking errors and high applied wrench decrease both $\eta$ and $\mu$ quickly. Thus, it returns to the demonstration mode and starts a new learning cycle. This process repeats for skills $R_1$ and $R_2$, respectively. 

The results of the experiment demonstrate the system’s capacity to acquire and generalize the rehabilitation skill across all directions with only a few demonstrations.

\subsection{Sensitivity Analysis}

For robust online learning and adaptation, convergence of the DMP weights is essential.
During the demonstration phase, the DMP updates its weights to track the demonstration.
However, with unsuitable parameter choices, the weights may oscillate over time instead of converging to a fixed value.
Such oscillations can ultimately produce an incorrect DMP weight once full autonomous mode is reached, causing a local mismatch between the demonstrations and the DMP reproductions. To identify the critical hyperparameters and determine appropriate tuning strategy for stable experimental results, we perform a sensitivity analysis on key hyper-parameters.  Specifically, we examine the forgetting factor $\lambda_{fg}$ in the Recursive Least Squares (RLS) algorithm~\cite{Gams2009} and the width $h$ of the von~Mises basis functions in \cref{eq: forcing}, as these influence both DMP's weight convergence and steady-state error between demonstrations and reproductions. 

To isolate these effects from hardware variability, we apply the learning framework to a one-dimensional toy sinusoidal signal in simulation, $0.45 + 0.05 \sin\!\bigl(2 \cdot 2\pi \cdot 0.3\, t\bigr),$ which resembles the $R_1$ exercises and has a duration of 40 s. For more complicated signals, we can follow similar tuning guidance.

We choose the number of basis functions as 30. An artificial learning level $\mu$ rises from zero at $t=28$ s to one at $t=30$ s to mimic the autonomy transition observed in real robot experiments. To evaluate DMP weight convergence, we record the DMP weights from $t=20$ s to $t=28$ s and compute their standard deviation (reported as “Weight std.”). The steady-state error is defined as the root-mean-square (RMS) difference between the DMP reproduced trajectory and the demonstration from $t=30$ s to $t=40$ s, after full autonomy is reached.

A trade-off emerges between DMP weight convergence, measured by the weight standard deviation, and the steady-state error. We vary the basis-function width as $h \in \{1, 3, 8, 31, 100\}$, corresponding to overlaps with approximately $6$, $3$, $2$, $1$, and $0.5$ neighboring basis functions, respectively.  As shown in \cref{tab:width}, larger widths (i.e., less overlap) improve weight convergence but increase the steady-state error. Balancing these effects, we select $h = 31$.

Similarly, \cref{tab:forget} demonstrates that a higher forgetting factor $\lambda_{fg}$ improves weight convergence but prolongs the learning phase.  
Excessively low or high values of $\lambda_{fg}$ increase the steady-state error.  
Considering both criteria, we select the intermediate value $\lambda_{fg}=0.9995$. 

\begin{table}[t]
\centering
\caption{Effect of radial basis function width $h$ on DMP weight convergence and steady-state error ($\lambda_{fg}=0.9995$).}
\label{tab:width}
\begin{tabular}{lccccc}
\hline
$h$ & 1 & 3 & 8 & 31 & 100 \\ \hline
Weight std.      & 17.93 & 8.44 & 4.56 & 3.58 & \textbf{3.47} \\
Mean RMS error   & 2.75  & \textbf{2.18} & 2.20 & 2.25 & 2.27 \\ \hline
\end{tabular}
\end{table}

\begin{table}[t]
\centering
\caption{Effect of RLS forgetting factor $\lambda_{fg}$ on DMP weight convergence and steady-state error ($h=31$).}
\label{tab:forget}
\begin{tabular}{lccccc}
\hline
$\lambda_{fg}$ & 0.99 & 0.995 & 0.999 & 0.9995 & 0.9999 \\ \hline
Weight std.      & 17.45 & 11.75 & 6.23 & 3.58 & \textbf{0.95} \\
Mean RMS error   & 2.69  & 2.49  & \textbf{2.24} & 2.25 & 2.38 \\ \hline
\end{tabular}
\end{table}

\section{CONCLUSIONS}

This work develops a tele-teaching framework to enable real-time learning and adaptation of the rehabilitation skill from remote therapist demonstration in $\mathbb{R}^3 \times \mathit{S}^3$ space. Rehabilitation skills, including translational and rotational trajectory, demonstrated through the \textit{Therapist-side Robot} are seamlessly transferred online to the \textit{Patient-side Robot} located remotely. The tele-teaching framework demonstrates remarkable speed and efficiency in adapting to new rehabilitation skills. The adaptation occurs smoothly within just a few iterations. 
The autonomy allocation is based on the learning level and the operator's intentions. This strategy not only ensures smooth transitions between human demonstration and autonomous robot execution but also empowers the user to dictate the cessation or initiation of teaching phases. 

Future works include extending to learn the interaction force and motion simultaneously in 6D space, and exploring the assist-as-needed rehabilitation strategy after a rehabilitation skill is learnt.

\addtolength{\textheight}{-12cm}   








\bibliographystyle{IEEEtran}
\bibliography{ref}

\end{document}